\documentclass[11pt,a4paper]{article}
\usepackage[hyperref]{acl2021}
\usepackage{times}
\usepackage{latexsym}

\usepackage{microtype}

\aclfinalcopy %
\def\aclpaperid{2877} %

\usepackage{graphicx}
\usepackage{booktabs}
\usepackage{multirow, multicol}

\usepackage[normalem]{ulem}
\useunder{\uline}{\ul}{}

\title{
It's All in the Heads: Using Attention Heads as a Baseline\\ for Cross-Lingual Transfer in Commonsense Reasoning
}

\author{Alexey Tikhonov\Thanks{     Equal contribution.}\\
  Yandex \\
  Berlin, Germany \\
  \texttt{altsoph@gmail.com} \\\And
  Max Ryabinin\footnotemark[1] \\
  Yandex, HSE University \\
  Moscow, Russia \\
  \texttt{mryabinin0@gmail.com} \\}

\date{}

\begin{document}
\maketitle
\begin{abstract}
Commonsense reasoning is one of the key problems in natural language processing, but the relative scarcity of labeled data holds back the progress for languages other than English.
Pretrained cross-lingual models are a source of powerful language-agnostic representations, yet their inherent reasoning capabilities are still actively studied.
In this work, we design a simple approach to commonsense reasoning which trains a linear classifier with weights of multi-head attention as features.
To evaluate this approach, we create a multilingual Winograd Schema corpus by processing several datasets from prior work within a standardized pipeline and measure cross-lingual generalization ability in terms of out-of-sample performance.
The method performs competitively with recent supervised and unsupervised approaches for commonsense reasoning, even when
applied to other languages in a zero-shot manner.
Also, we demonstrate that most of the performance is given by the same small subset of attention heads for all studied languages, which provides evidence of universal reasoning capabilities in multilingual encoders.
\end{abstract}

\section{Introduction}

Neural networks have achieved remarkable progress in numerous tasks involving natural language, such as machine translation \cite{bahdanau2014neural,kaplan2020scaling,arivazhagan2019massively}, language modeling \cite{gpt3}, open-domain dialog systems \cite{meena,blender}, and general-purpose language understanding \cite{devlin-etal-2019-bert, deberta}. 
However, the fundamental problem of commonsense reasoning has proven to be quite challenging for modern methods and arguably remains unsolved up to this day.
The tasks that aim to measure reasoning capabilities, such as the Winograd Schema Challenge \cite{levesque2012winograd}, are deliberately designed not to be easily solved by statistical approaches, which are a foundation of most deep learning methods. Instead, these tasks require implicit knowledge about properties of real-world entities and their relations in order to resolve inherent ambiguities of natural language.

Figure \ref{fig:winograd-example} illustrates the gist of this task: given a sentence and a pronoun (\underline{they}), the goal is to choose the word that this pronoun refers to from two options (\textit{The town councilors} or \textit{the demonstrators}). While picking the right answer is straightforward for humans, the lack of explicit clues makes it hard for machine learning algorithms to perform better than majority vote or random choice.

\begin{figure}
    \textit{The town councilors} refused to give \textit{the demonstrators} a permit because \underline{they} feared violence.\\
    \textbf{Answer:} The town councilors
    \caption{Example of a Winograd Schema problem. The resolved pronoun is underlined, two options are highlighted with an italic font.}
    \vspace{-22pt}
    \label{fig:winograd-example}
\end{figure}

Recently large Transformer-based masked language models (MLMs) \cite{devlin-etal-2019-bert} were shown to achieve impressive results on several benchmark datasets for commonsense reasoning \cite{sakaguchi2020winogrande,kocijan-etal-2019-surprisingly,klein-nabi-2020-contrastive}. However, the best-performing methods frequently involve finetuning the entire model on large enough corpora with varying degrees of supervision; apart from providing initial parameter values, the pretrained trained language model is not used for predictions.

Moreover, these methods have mostly been evaluated on English language datasets, despite increasing interest in multilingual evaluation for NLP \cite{hu2020xtreme} and the existence of multilingual encoders \cite{conneau-etal-2020-unsupervised,lample2019crosslingual}. The XCOPA dataset \cite{ponti-etal-2020-xcopa} was recently proposed as a benchmark for multilingual commonsense reasoning, yet its task is different from the pronoun resolution problem described above. Versions of Winograd Schema Challenge exist in different languages, but each version comes with slight differences in task specification. This makes holistic cross-lingual evaluation of new commonsense reasoning approaches a quite difficult problem for researchers in the area.

In this work, we propose a simple supervised method for commonsense reasoning, which trains a linear classifier on the self-attention weights between the pronoun and two answer options. 
To evaluate our method and facilitate research in multilingual commonsense reasoning, we aggregate existing Winograd Schema datasets in English, French, Japanese, Russian, Portuguese, and Chinese languages, converting them to a single format with a strict task definition. Our approach performs comparably to supervised and unsupervised baselines in this setting with both multilingual BERT and XLM-R models as backbone encoders.

Moreover, we find that the same set of attention heads can be used to solve reasoning tasks in all languages, which hints at the emergence of language-independent linguistic functions in cross-lingual models and supports the conclusions made by prior work~\cite{chi-etal-2020-finding,li-etal-2020-heads}. Interestingly, when using an unsupervised attention-based method~\cite{klein-nabi-2019-attention}, we observe that restricting the choice of heads to this set also improves the results of this baseline. 
This result suggests that the key to improved performance of such approaches might lie in the right choice of heads rather then the exact attention values.

To summarize, our contributions are as follows:
\vspace{-4pt}
\begin{itemize}
\setlength{\itemsep}{0pt}
    \item We offer a simple supervised method to utilize self-attention heads of pretrained language models for commonsense reasoning.
    \item We compile XWINO --- a dataset of Winograd schemas in six languages, bringing all tasks to the same format\footnote{The datasets and code are available at \href{https://github.com/yandex-research/crosslingual_winograd}{\texttt{github.com/\\yandex-research/crosslingual\_winograd}}}. When evaluated on this dataset, our method performs competitively to strong baselines from prior work.
    \item We demonstrate that in cross-lingual models, there exists a small subset of attention heads specializing in \textit{universal} commonsense reasoning. This reveals new linguistic properties of masked language models trained on multiple languages.
\end{itemize}

\section{Related work}
\subsection{Winograd Schema challenges}

The Winograd Schema Challenge (WSC) was proposed as a challenging yet practical benchmark for evaluation of machine commonsense reasoning~\cite{levesque2012winograd}. Since its introduction, several English-language benchmarks of varying difficulty and size were also proposed: notable examples include Definite Pronoun Resolution~\cite{rahman-ng-2012-resolving} and Pronoun Disambiguation Problem~\cite{morgenstern2016planning} datasets, as well as WinoGrande, which consists of 44k crowdsourced examples~\cite{sakaguchi2020winogrande}. A version of WSC is also included in the popular SuperGLUE language understanding benchmark~\cite{wang2019superglue}, where it is reformulated as a natural language inference problem.

There also exist variations of WSC in other languages: French~\cite{amsili-seminck-2017-google}, Japanese~\cite{shibata2015winograd}, Russian~\cite{shavrina-etal-2020-russiansuperglue}, Portuguese~\cite{melo2019winograd}, and Chinese~\cite{bernard-han-2020-mandarinograd}. We use these datasets in our study to create a multilingual dataset for commonsense reasoning.

Although in general the task definition of Winograd Schema Challenge was formalized to some degree, both succeeding datasets and methods proposed by users of these datasets have introduced various changes to the task specification and even the input format. In particular, a work by~\citet{liu-etal-2020-precise} provides a thorough comparison of different ways to formalize the task for WSC and shows that the same model can give widely varying results depending on the evaluation framework. We describe our efforts to convert different datasets to a single format in Section \ref{sect:dataset}.

\subsection{Language models applied to commonsense reasoning}
Several works attempt to solve Winograd Schema Challenge by utilizing pretrained language models. For example, \citet{trinh2018simple} propose to rank possible answers with an ensemble of RNN language models by substituting the pronoun with each of the options. Recently, \citet{klein-nabi-2019-attention} introduced Maximum Attention Score (MAS) for commonsense reasoning. This method uses the outputs of multi-head attention from each layer and scores each candidate answer based on the number of heads for which this answer has the highest attention value. We use the first (adapted to masked language models as proposed by~\citealp{salazar-etal-2020-masked}) and the second approaches as baselines in the experiments. In essence, our method can be compared to MAS, but as we demonstrate in Section~\ref{sect:experiments}, several algorithm design differences along with task supervision allow us to significantly improve the commonsense reasoning performance.

Large pretrained Transformer models, such as BERT \cite{devlin-etal-2019-bert}, have also enabled rapid progress of supervised methods for WSC. One such method is given by~\citet{sakaguchi2020winogrande}: the authors propose to concatenate the sentence and one of the options and to use the [CLS] token representation of the resulting sequence for binary classification. Also, \citet{kocijan-etal-2019-surprisingly} propose a margin-based loss function which aims to increase the log-probability of the correct answer as a replacement for the masked pronoun. We evaluate these methods in our experiments without training on large in-domain datasets; as we show, both methods are prone to overfitting when applied to several hundreds of examples.

\subsection{Cross-lingual encoder models}

Multilingual representations have been a long-standing goal of the research community: they allow to serve fewer models for a wide range of languages and to improve the results on low-resource languages. \citet{ruder2019survey} gives a detailed survey of different cross-lingual word embedding approaches, as well as the history of cross-lingual representations in general. 

In this work, we are interested in the latest developments in multilingual Transformer masked language models~\cite{devlin-etal-2019-bert,lample2019crosslingual, conneau-etal-2020-unsupervised,siddhant-etal-2020-leveraging} that were driven by the advances in transfer learning for NLP \cite{howard2018universal,devlin-etal-2019-bert}. In particular, we use pretrained multilingual BERT (mBERT,~\citealp{devlin-etal-2019-bert}) and XLM-RoBERTa (XLM-R,~\citealp{conneau-etal-2020-unsupervised}) for all our experiments.

Recently, there has been increasing interest in the evaluation of multilingual models: as a result, several benchmarks, including XTREME~\cite{hu2020xtreme}, XNLI~\cite{conneau-etal-2018-xnli} and XCOPA~\cite{ponti-etal-2020-xcopa} were introduced. Although XCOPA is a commonsense reasoning dataset, it is meant to serve as a multilingual version of the COPA dataset~\cite{roemmele2011choice}, which offers a problem different from pronoun resolution. In this work, we aimed to create a multilingual counterpart of more widely used Winograd Schema Challenge, so that any future methods for commonsense reasoning can be easily evaluated on languages other than English.

\subsection{Functions of Transformer heads}

Previous works have demonstrated that it is possible to perform unsupervised zero-shot consistency parsing with attention heads of pretrained cross-lingual models~\cite{kim2020chartbased, li-etal-2020-heads}. In our work, we extend these findings to a conceptually different task of commonsense reasoning. This task has significant overlap with coreference resolution, which was shown to be encoded in specific heads of monolingual BERT~\cite{clark-etal-2019-bert,tenney-etal-2019-bert}.

Motivated by similar results for monolingual models, several works have previously demonstrated that models such as multilingual BERT encode grammatical relations~\cite{chi-etal-2020-finding} and can perform zero-shot entity recognition, as well as POS-tagging~\cite{pires-etal-2019-multilingual}. Besides presenting evidence for universality in pronoun resolution, which was not studied before, our analysis relies on attention heads instead of extracting representations from intermediate layer outputs.

\section{Common sense from attention}

In this section, we first give a formal definition of the commonsense reasoning task, most commonly encountered in Winograd Schema Challenge and its successors. Then, we provide necessary background information about the Transformer architecture for transfer learning and describe our proposed solution for this task.

\subsection{Exact task specification}

It is known that commonsense reasoning performance can vary greatly due to changes in task formulation: for example, recent work by \citet{liu-etal-2020-precise} reports improvements of up to 6 points when posing the task as multiple choice instead of binary classification. Thus, as per recommendations from this work and in order to create a unified dataset, we choose the definition of the Winograd Schema problem which is as strict as possible. 

The definition is as follows: the system receives a sentence with a pronoun and has to choose the noun (or noun phrase) that this pronoun refers to. For this choice, the system has two options; both of which, along with the pronoun, are always included as substrings of the initial sentence. We intentionally do not restrict the choice of sentence representation or the framing of the task in order to evaluate a diverse range of solutions.

Although the requirements listed above are quite general and intuitive when working with WSC, some of the datasets we employ have samples that do not conform to them. 
For example, it might be the case that the pronoun occurs at several positions in the sentence without explicit indication of the one to be resolved.
For all such examples, we attempt to convert them to standardized instances by hand and drop them only if it is not possible via simple means: otherwise, the right answer to the problem is misspecified. We give a detailed description of our solution in Section \ref{sect:filtering}.

\subsection{Transformers for sentence representations}

Our method heavily relies on the specifics of the Transformer architecture~\cite{transformer}, which has attracted increased interest in NLP recently due to its generation~\cite{t5,gpt3} and transfer learning~\cite{devlin-etal-2019-bert,roberta} capabilities. 

This architecture consists of several sequential layers, where each layer contains a feed-forward block and a self-attention block. Inside the self-attention block, there are multiple \textit{attention heads}: each head first linearly projects the input sequence $z=[z_1,\ldots,z_i,\ldots,z_n]$ into sequences of queries $q_i$, keys $k_i$ and values $v_i$, then computes the attention weights as softmax-normalized values of pairwise dot products between all keys and all queries:
\begin{equation}
\label{ref:eqn:attention}
    \alpha_{ij}=\frac{\exp(q_i^T k_j)}{\sum_{l=1}^n \exp(q_i^T k_l)}
\end{equation}

These weights are then used to combine the values into a single vector for each input vector, and the layer output is a linear combination of all attention head outputs.

\subsection{Our approach}
\label{sect:approach}
The method proposed in this work uses intermediate outputs of a Transformer masked language model with $L$ layers and $H$ heads in each layer. Given an instance of the Winograd Schema problem, we take the input sentence and mask the pronoun that needs to be resolved. After that, we feed the resulting sentence to the language model and obtain the activations of each self-attention layer as a tensor $L\times H\times T$, where $T$ is the number of tokens that constitute the candidate answer. Here, we can either take the attention from the pronoun to the candidate or vice versa. 

After aggregating the attention outputs by computing the mean or the maximum over $T$, we have two matrices for each of two possible answers, which are then flattened into vectors. Combining these vectors, we obtain an input for the binary classification task with class 0 corresponding to the first answer being correct and class 1 corresponds to the second one. Given a dataset of such inputs, we can train a logistic regression to predict the class from the multi-head attention weights $\alpha$.

There are several design choices which define the exact implementation of our method. We describe them below; for each design choice, we underline the best-performing option as found by the ablation study in Section \ref{sect:ablation}.
\paragraph{Feature combination:} With two feature vectors for candidate answers, we can either concatenate them or \underline{subtract} the vector of the second candidate from the vector of the first one.
\paragraph{Pooling over tokens:} As the candidates can have different length, we need to transform the attention outputs to feature vectors of the same size. This can be done by one of two simple forms of aggregation: \underline{mean-} or max-pooling.
\paragraph{Attention direction:} Observe from Equation~\ref{ref:eqn:attention} that in general, $\alpha_{ij}\neq\alpha_{ji}$. To find the optimal configuration, we evaluate both options of either attending to the \underline{candidate} or the pronoun.

\section{Dataset}
\label{sect:dataset}

In this section, we describe our procedure of building XWINO --- a multilingual commonsense reasoning benchmark using Winograd Schema Challenge problems.
We create it by combining several monolingual collections for six languages, each described in previously published works. 

We intentionally do not use XCOPA~\cite{ponti-etal-2020-xcopa} as it is aimed at a different problem: instead of operating at the word level, the task of this dataset is to connect the premise and one of two hypotheses, both of which are complete sentences. Because direct application of attention-based reasoning to sentence-level tasks is a non-trivial research question, we leave it to future work.

\subsection{Languages}

For the English language, we work with the data from the original WSC task\footnote{Specifically, the WSC285 version.}~\cite{levesque2012winograd}, as well as the SuperGLUE benchmark~\cite{wang2019superglue} and the Definite Pronoun Resolution dataset~\cite{rahman-ng-2012-resolving}. For French and Japanese, we use datasets published by \citet{amsili-seminck-2017-google} and \citet{shibata2015winograd} respectively. We also include the corresponding part from the Russian SuperGLUE benchmark~\cite{shavrina-etal-2020-russiansuperglue}, a collection of Winograd Schemas in Chinese from the WSC website\footnote{\url{https://cs.nyu.edu/faculty/davise/papers/WinogradSchemas/WSChinese.html}}, and the Portuguese version of WSC~\cite{melo2019winograd} into our multilingual benchmark.

In addition, we attempted to use Mandarinograd~\cite{bernard-han-2020-mandarinograd} --- a Mandarin Chinese version of WSC. However, this dataset contains questions instead of pronouns that need to be resolved. As such, we were unable to incorporate its contents without significantly changing the task.

\subsection{Preprocessing and filtering}
\label{sect:filtering}
As the datasets for different languages were released in several different formats, in order to have a unified evaluation framework, we needed to convert them all to the same schema. Unfortunately, due to the differences in task formalization we were unable to convert certain examples without completely changing them; as a result, these examples had to be removed from the dataset. Still, our main priority was to maintain the same task format while keeping as many examples as possible; to this end, we fixed minor annotation inconsistencies by hand wherever possible. 

Below we describe the steps of our pipeline. First, several examples had more than two candidate choices, i.e. more than one incorrect option is given. We convert these examples into several binary choice problems and report the original dataset sizes after executing this step. Next, the main issue we faced was that the right answer is not included as a substring of the input sentence. Often this can be explained by missing articles, typos or differences in word capitalization. We attempt to fix all such errors in these cases.

The resulting dataset sizes are listed in Table~\ref{tab:dataset-sizes}; it can be seen that our conversion pipeline discards approximately 29\% of data. In the future, more effort could be directed towards constructing a linguistically diverse, large-scale and balanced multilingual Winograd Schema dataset. Yet, as shown in Section~\ref{sect:experiments}, XWINO already allows us to distinguish recent commonsense reasoning models by their performance.

\begin{table}[t]
\centering
\begin{tabular}{@{}lccc@{}}
\toprule
Language   & Before & After & Remaining, \% \\ \midrule
English   &    2605    &   2325 & 89.25   \\
French    &   214     &   83  &  38.79 \\
Japanese   &   1886     &   959  & 50.85  \\
Russian   &    569    &    315 & 55.36  \\
Chinese     &    18    &    16 & 92.28  \\
Portuguese & 285 & 263& 88.89 \\\midrule
Total & 5577 & 3961 & 71.02\\
\bottomrule
\end{tabular}
\caption{Dataset sizes before and after filtering.}
\label{tab:dataset-sizes}
\vspace{-16pt}
\end{table}

\section{Experiments}
\label{sect:experiments}
Below we describe the experimental setup used to evaluate cross-lingual transfer capabilities of different approaches to commonsense reasoning and report the results.
Note that we also aim to study the universal reasoning properties of attention heads, and thus we do not evaluate our method on common monolingual Winograd Schema datasets.

\begin{table*}[t]
\centering
\small
\begin{tabular}{@{}llccccccc@{}}
\toprule
Model & Train lang & en & fr & ja & ru & zh & pt & Avg \\ \midrule
\multicolumn{9}{c}{Unsupervised} \\ \midrule
MLM prob. ranking & - & 53.6 & 53.0 & 52.5 & 51.8 & 31.3 & 50.2 & \bf 52.8 \\
Pseudo-perplexity & - & 53.0 & 54.2 & 49.5 & 53.7 & 56.3 & 49.4 & 52.0 \\
MAS & - & 52.3 & 51.8 & 50.2 & 52.7 & 56.3 & 49.1 & 51.6 \\ \midrule
\multicolumn{9}{c}{Supervised} \\ \midrule
\multirow{6}{*}{\citet{kocijan-etal-2019-surprisingly}} & en & - & 52.5±4.1 & 51.4±0.8 & 51.2±1.1 & 48.8±9.3 & 51.2±1.5 & 51.0 \\
 & fr & 50.9±0.4 & - & 51.5±0.7 & 51.3±0.9 & 56.2±9.9 & 49.2±1.2 & 51.8 \\
 & ja & 51.0±0.7 & 50.8±2.2 & - & 50.1±1.0 & 55.0±6.8 & 49.9±1.1 & 51.4 \\
 & ru & 50.7±0.5 & 51.3±2.0 & 51.7±1.0 & - & 51.2±10.3 & 51.0±0.3 & 51.2 \\
 & zh & 50.9±0.2 & 50.1±1.6 & 50.8±0.4 & 50.5±0.0 & - & 53.0±1.4 & \bf 51.1 \\
 & pt & 51.1±0.5 & 54.0±3.9 & 51.3±0.6 & 49.3±0.5 & 53.8±7.1 & - & 51.9 \\ \midrule
\multirow{6}{*}{Ours} & en & - & 53.7±1.6 & 52.2±0.4 & 60.1±0.4 & 51.2±4.7 & 53.9±0.4 & \bf 54.2 \\
 & fr & 51.7±1.1 & - & 51.1±1.1 & 52.2±2.6 & 53.8±3.1 & 50.9±1.1 & \bf 51.9 \\
 & ja & 52.7±0.3 & 55.4±0.8 & - & 58.0±1.2 & 50.0±4.0 & 51.3±1.3 & \bf 53.5 \\
 & ru & 55.5±0.4 & 52.3±2.7 & 52.3±0.3 & - & 52.5±5.0 & 52.0±0.7 & \bf 52.9 \\
 & zh & 49.8±2.0 & 48.2±4.5 & 50.5±1.4 & 50.3±4.9 & - & 49.0±1.4 & 49.6 \\
 & pt & 54.7±0.5 & 52.8±3.2 & 51.7±0.6 & 57.7±1.2 & 50.0±4.0 & - & \bf 53.4 \\ \bottomrule
\end{tabular}
\caption{Results for multilingual BERT, best result is denoted by bold font.}
\label{tab:mbert-results}
\vspace{-5pt}
\end{table*}

\subsection{Setup}
\paragraph{Models} We use multilingual BERT~\cite{devlin-etal-2019-bert} and XLM-R-Large~\cite{conneau-etal-2020-unsupervised}, as these models are frequently used in other multilingual evaluation literature. The first model has 12 layers with 12 attention heads each, whereas the second model is a 24-layer Transformer with 16 attention heads on each layer.
We do not evaluate XLM-R-Base or multilingual translation encoders~\cite{siddhant-etal-2020-leveraging} because we take two best-performing models according to the XTREME benchmark~\cite{hu2020xtreme}.

For our method, we use an implementation of logistic regression from scikit-learn~\cite{scikit-learn} with default hyperparameters as a linear classifier over attention weights.

\paragraph{Evaluation} For unsupervised methods, we directly apply each method to each language subset and report the classification accuracy. For supervised methods, we first choose a single language for training and generate random train-validation-test splits, leaving 10\% of data both for validation and testing subsets. For each language, we create 5 random train-validation splits to estimate the standard deviation of metrics, while keeping the same test set to keep the results comparable. Additionally, we test each trained model for a language on all other languages in a zero-shot setting, reporting averaged performance as well.

\subsection{Baselines}
To compare our approach with currently popular methods, we also evaluate a wide set of well-performing approaches described in earlier works:

\paragraph{Unsupervised}
We use three entirely unsupervised baselines inspired by prior work. For the first approach, we replace the pronoun by the number of \texttt{[MASK]} tokens equal to the length of each candidate answer and compare the MLM probabilities. For the second approach, we replace the pronoun with each of the answers and rank the candidates by ``pseudo-perplexity''~\cite{salazar-etal-2020-masked}, inspired by the results of \citet{trinh2018simple}.
Both baselines use normalized scores with respect to the candidate word length. 

The third unsupervised baseline is Masked Attention Score (MAS), described in \citet{klein-nabi-2019-attention}. Similarly to our method, this approach relies on attention weights for prediction; however, they are utilized differently and the model is unable to discover an optimal subset of heads.

\paragraph{Supervised} First, we evaluated the masked language model finetuning approach suggested by the authors of WinoGrande~\cite{sakaguchi2020winogrande}.
However, in our experiments there are no additional large-scale datasets; we found that with reference hyperparameters, the authors' implementation quickly overfits the training data for all languages in our relatively small benchmark, achieving less than 50\% zero-shot accuracy on average.

In addition, we used the margin-based classification approach described in~\cite{kocijan-etal-2019-surprisingly}. This method achieves competitive results and outperforms unsupervised baselines in most setups, so we include it in our comparison.

\subsection{Results}

The results of our experiments for multilinual BERT and XLM-R-Large are shown in Tables~\ref{tab:mbert-results} and~\ref{tab:xlmr-results} respectively. It can be seen that despite using only the attention weights as features, our method can outperform unsupervised approaches and performs competitively with a state-of-the-art supervised approach in several setups.
Notably, the quality improves significantly when going from BERT to XLM-R: this goes in line with previous work on evaluation of cross-lingual encoders~\cite{hu2020xtreme}. At the same time, the quality of our method improves more significantly than of that suggested by~\citet{kocijan-etal-2019-surprisingly}: this may be explained by a greater parameter count and a higher number of attention heads with more distinct specializations.

\begin{table*}[t!]
\centering
\small
\begin{tabular}{@{}llccccccc@{}}
\toprule
Model & Train lang & en & fr & ja & ru & zh & pt & Avg \\ \midrule
\multicolumn{9}{c}{Unsupervised} \\ \midrule
MLM prob. ranking & - & 58.8 & 56.6 & 61.7 & 57.5 & 56.3 & 56.7 & \bf 59.2 \\
Pseudo-perplexity & - & 58.5 & 54.2 & 58.2 & 59.7 & 56.3 & 54.8 & 58.2 \\
MAS & - & 57.2 & 56.6 & 53.9 & 58.1 & 50.0 & 53.6 & 56.2 \\ \midrule
\multicolumn{9}{c}{Supervised} \\ \midrule
\multirow{6}{*}{\citet{kocijan-etal-2019-surprisingly}} & en & - & 70.6±2.0 & 81.4±1.8 & 74.8±1.1 & 72.5±5.6 & 74.3±1.2 & \bf 74.7 \\
 & fr & 59.6±0.5 & - & 65.8±0.2 & 58.9±0.3 & 56.2±0.0 & 56.7±0.9 & 59.4 \\
 & ja & 70.4±3.8 & 62.7±2.7 & - & 66.1±1.9 & 63.7±5.2 & 63.7±2.7 & 65.3 \\
 & ru & 67.1±4.8 & 62.2±4.2 & 69.2±3.2 & - & 56.2±0.0 & 61.0±4.3 & 63.1 \\
 & zh & 59.2±0.0 & 57.8±0.0 & 65.5±0.0 & 58.7±0.0 & - & 56.3±0.0 & \bf 59.5 \\
 & pt & 67.1±3.6 & 61.2±2.2 & 69.4±3.1 & 65.6±3.7 & 60.0±3.4 & - & 64.6 \\ \midrule
\multirow{6}{*}{Ours} & en & - & 67.5±1.3 & 69.1±0.2 & 70.4±0.5 & 60.0±3.1 & 66.8±0.9 & 66.7 \\
 & fr & 66.1±0.7 & - & 63.3±0.8 & 67.0±1.2 & 60.0±5.0 & 61.4±1.2 &\bf 63.6 \\
 & ja & 70.1±0.4 & 67.2±1.4 & - & 72.4±0.6 & 61.3±2.5 & 65.9±0.8 &\bf 67.4 \\
 & ru & 68.7±0.5 & 65.8±2.7 & 65.7±0.5 & - & 63.7±4.7 & 64.4±0.8 &\bf 65.7 \\
 & zh & 51.1±8.5 & 48.2±8.6 & 52.4±8.8 & 51.3±10.8 & - & 50.0±8.1 & 50.6 \\
 & pt & 68.9±0.6 & 67.0±1.6 & 68.1±0.4 & 69.5±0.2 & 63.7±4.7 & - &\bf 67.4 \\ \bottomrule
\end{tabular}
\caption{Results for XLM-R-Large, best result is denoted by bold font.}
\label{tab:xlmr-results}
\vspace{-5pt}
\end{table*}

\subsection{Ablation study}
\label{sect:ablation}
Here we compare several algorithm versions listed in Section~\ref{sect:approach}. 
We train all models on the English part of XWINO and evaluate on all other languages, using validation subset performance as our target metric. 
As the Table~\ref{tab:ablation} demonstrates, each choice leads to drops in performance, with the most influential being the choice of feature concatenation instead of taking the difference and attention direction being the least important decision.

\begin{table}[t]
\small
\centering
\setlength{\tabcolsep}{2pt}
\begin{tabular}{@{}lccccccc@{}}
\toprule
Method & Valid & fr & ja & ru & zh & pt & Avg \\ \midrule
Ours (Section~\ref{sect:approach}) &\bf 55.4 & 53.7 & 52.2 &\bf 60.1 & 51.2 &\bf 53.9 &\bf 54.2 \\
Concat & 53.0 &\bf 54.9 &\bf 52.3 & 56.3 &\bf 53.8 &\bf 53.9 &\bf 54.2 \\
Max pooling & 53.6 &\bf 52.3 & 52.3 & 59.9 & 50.0 & 51.6 & 53.2 \\
Attn from pronoun & 54.8 & 53.0 & 52.2 & 57.4 & 47.5 & 52.9 & 52.6 \\ \bottomrule
\end{tabular}
\normalsize
\caption{Ablation study results for models trained on the English subset; the best result is in bold.}
\label{tab:ablation}
\vspace{-10pt}
\end{table}

\section{Analyzing the attention heads}

In this section, we intend to analyze the reasons behind competitive generalization performance of our approach. 
It is known what individual attention heads sometimes play consistent and often linguistically-interpretable roles \cite{voita1}. 
We compare the subsets of heads learned on different languages and measure their impact on the prediction quality.

\subsection{Universal commonsense reasoning}

\begin{figure}[t]
    \centering
    \includegraphics[width=\columnwidth]{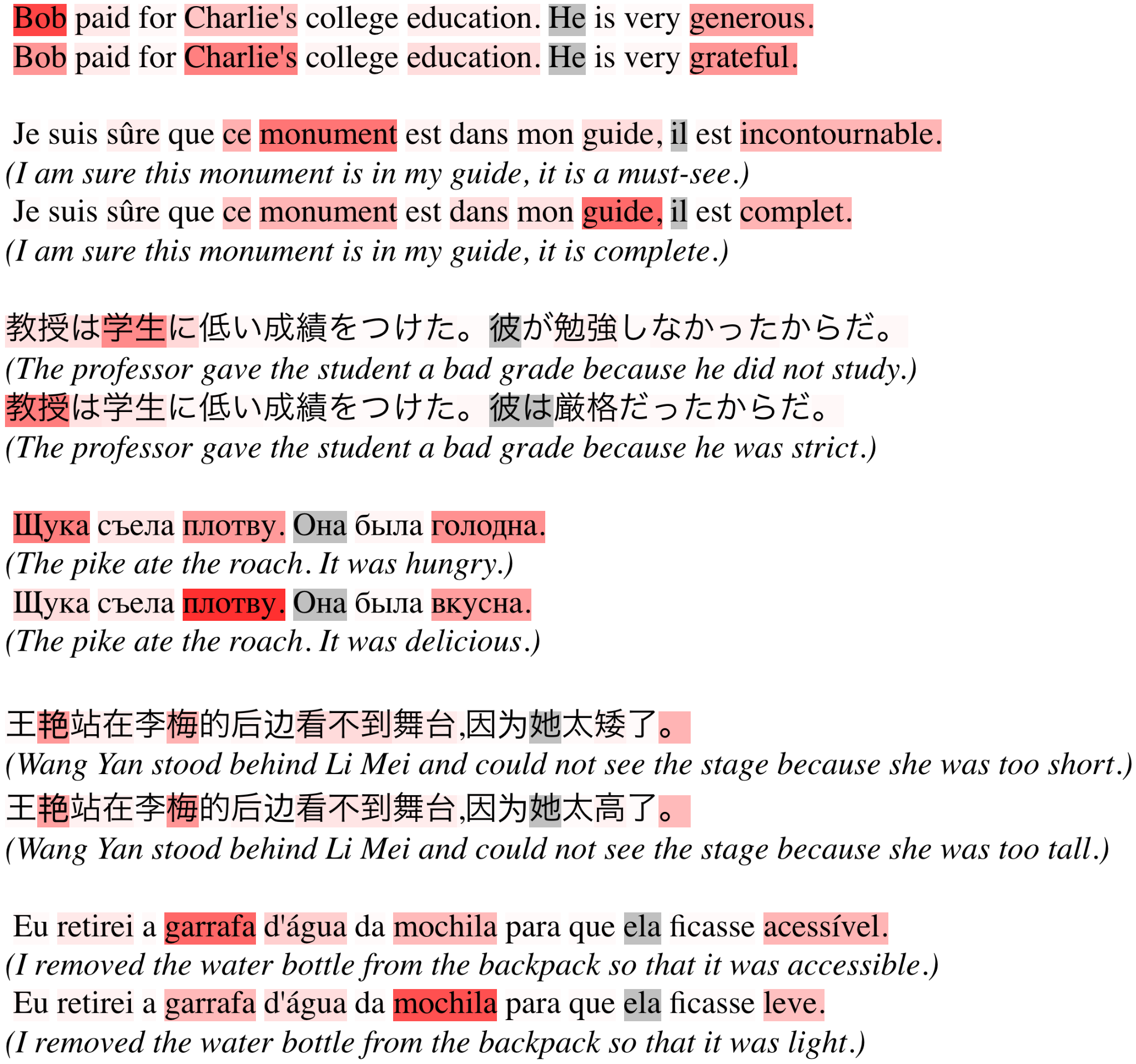}
    \caption{Averaged attention from the pronoun when using top-5 common heads.}
    \label{fig:example-attn}
    \vspace{-10pt}
\end{figure}

For the first experiment, we rank the heads for models trained on all languages with the XLM-R\footnote{The results for mBERT are available in Appendix~\ref{app:analysis_mbert}.} representations by the absolute value of the weight. Then, we consider the top-5 heads which are ranked highest on average across all languages. These common heads are located in the higher layers of the model, which was shown previously to encode mainly semantic features~\cite{raganato-tiedemann-2018-analysis,jo-myaeng-2020-roles}, which intuitively corresponds to the tasks the model needs to solve for pronoun resolution. Figure~\ref{fig:example-attn} shows the average attention weights of these heads for each word in several example sentences.

\begin{table*}[t]
\small
\centering
\begin{tabular}{@{}llccccccc@{}}
\toprule
Train lang & Heads & en & fr & ja & ru & zh & pt & Avg \\ \midrule
\multicolumn{9}{c}{MAS (unsupervised)} \\ \midrule
\multicolumn{1}{l}{\multirow{3}{*}{-}} & All & 57.2 & 56.6 & 53.9 & 58.1 & 50.0 & 53.6 & 56.2 \\
\multicolumn{1}{c}{} & Random & 57.8 & 56.6 & 56.9 & 61.6 & 50.0 & 56.7 & 56.6 \\
\multicolumn{1}{c}{} & Common &\bf 65.8 &\bf 62.7 &\bf 64.9 &\bf 67.3 &\bf 68.8 &\bf 64.3 &\bf 65.6 \\ \midrule
\multicolumn{9}{c}{Ours (supervised)} \\ \midrule
\multirow{3}{*}{en} & All & \multirow{3}{*}{-} & 67.5 &\bf 69.1 &\bf 70.4 & 60.0 &\bf 66.8 &\bf 66.7 \\
 & Random &  & 62.0 & 64.4 & 67.4 & 60.0 & 65.4 & 63.9 \\
 & Common &  &\bf 68.4 & 66.6 & 68.5 & 62.5 & 65.3 & 66.3 \\ \cmidrule(l){2-9} 
\multirow{3}{*}{fr} & All & 66.1 & \multirow{3}{*}{-} & 63.3 &\bf 67.0 & 60.0 & 61.4 & 63.6 \\
 & Random & 59.9 &  & 58.3 & 60.7 & 58.8 & 57.2 & 59.0 \\
 & Common &\bf 66.7 &  &\bf 63.8 & 66.7 &\bf 63.7 &\bf 63.1 &\bf 64.8 \\ \cmidrule(l){2-9} 
\multirow{3}{*}{ja} & All &\bf 70.1 &\bf 67.2 & \multirow{3}{*}{-} &\bf 72.4 & 61.3 &\bf 65.9 &\bf 67.4 \\
 & Random & 66.0 & 62.2 &  & 68.0 & 59.4 & 65.3 & 64.2 \\
 & Common & 68.9 & 66.7 &  & 69.5 &\bf 62.5 & 64.9 & 66.5 \\ \cmidrule(l){2-9} 
\multirow{3}{*}{ru} & All &\bf 68.7 &\bf 65.8 & 65.7 & \multirow{3}{*}{-} &\bf 63.7 & 64.4 &\bf 65.7 \\
 & Random & 66.0 & 62.3 & 64.3 &  & 59.4 &\bf 64.6 & 63.3 \\
 & Common & 68.0 & 64.6 &\bf 66.5 &  &\bf 63.7 &\bf 64.6 & 65.5 \\ \cmidrule(l){2-9} 
\multirow{3}{*}{zh} & All & 51.1 & 48.2 & 52.4 & 51.3 & \multirow{3}{*}{-} & 50.0 & 50.6 \\
 & Random &\bf 59.4 &\bf 54.7 &\bf 58.6 &\bf 61.0 &  &\bf 58.0 &\bf 58.3 \\
 & Common & 46.4 & 47.2 & 49.4 & 46.8 &  & 46.9 & 47.4 \\ \cmidrule(l){2-9} 
\multirow{3}{*}{pt} & All &\bf 68.9 &\bf 67.0 &\bf 68.1 &\bf 69.5 &\bf 63.7 & \multirow{3}{*}{-} &\bf 67.4 \\
 & Random & 66.2 & 62.3 & 64.6 & 67.1 & 60.0 &  & 64.0 \\
 & Common & 67.9 & 65.5 & 66.0 & 68.2 &\bf 63.7 &  & 66.3 \\ \bottomrule
\end{tabular}
\normalsize
\caption{Performance of models trained with different subsets of XLM-R-Large attention heads.}
\label{tab:all-vs-common}
\end{table*}

After we locate the most important common heads, we train linear classifiers restricted to these heads as features only for every language. To evaluate the importance of head choice, we also provide the performance of linear classifiers trained on a fixed subset of 5 random heads. The results of this experiment can be seen in Table~\ref{tab:all-vs-common}; we observe that using the same top-5 heads (only $1.3\%$ of the total number) across all languages preserves or even improves the results. The only exception is Chinese, which might not have enough labeled data to extract a sufficient amount of task-specific information. It means that a very small subset of attention weights is required to perform commonsense reasoning in all evaluated languages. This further supports the previous results on the analysis of linguistic universals in cross-lingual models \cite{chi-etal-2020-finding,wang-etal-2019-cross}. 

Moreover, restricting the subset of heads used in the MAS baseline to those selected by the classifiers significantly improves the quality of this unsupervised method as well, nearly closing the gap with the results obtained with supervision. This leads us to the conclusion that initially the poor performance of MAS might be caused by the suboptimal choice of attention heads; when the right heads are selected, their weights do not impact the predictions as significantly. Future unsupervised methods for commonsense reasoning can use that information to pay more attention to the choice of heads, which is currently a less explored subject. 

\subsection{The impact of number of heads}

In this experiment, we directly study the connection between the number of heads and the quality of predictions. Specifically, after training a model with a full set of attention heads, we order them by the absolute value. Then, we retrain the model while keeping only the top-$N$ important heads.

\begin{figure}[t]
    \centering
    \includegraphics[width=\columnwidth]{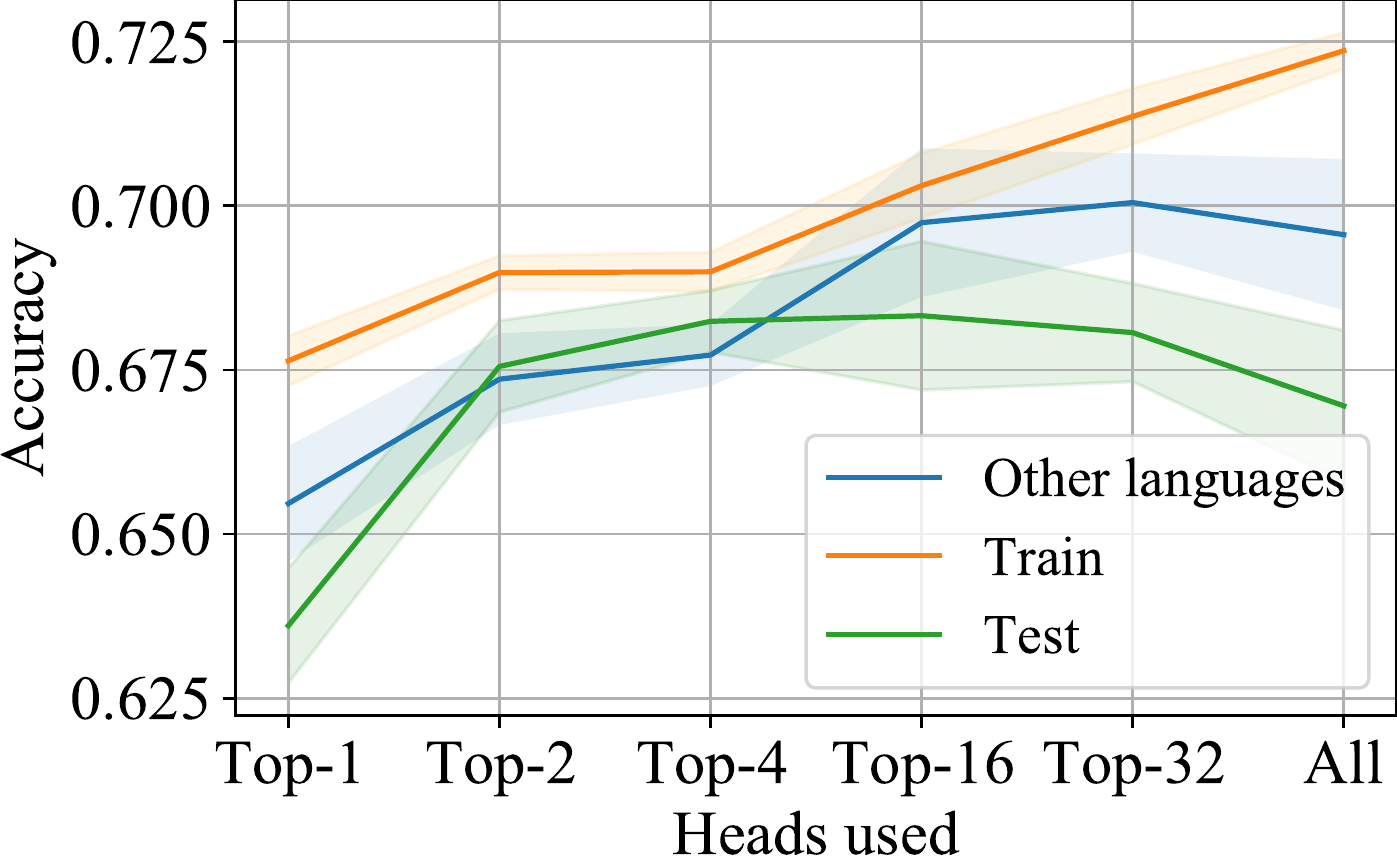}
    \caption{Effect of the number of XLM-R attention heads used when training on English data. Shaded areas show standard deviation across runs.}
    \label{fig:heads_en}
    \vspace{-10pt}
\end{figure}

Figure~\ref{fig:heads_en} displays the results of our study for the English language; results for other languages can be seen in Appendix~\ref{app:heads}. From these results, we find that although the training accuracy monotonically increases with the number of used attention weights, the optimal amount of heads for cross-lingual generalization is approximately equal to 16. This number is optimal or near-optimal for other languages as well, which might mean that as the number of features grows, the model either simply overfits the data or starts relying on features that are not universal for all languages.

\section{Conclusion}

In this work, we offer a simple supervised method to utilize pretrained language models for commonsense reasoning. It relies only on the outputs of self-attention and outperforms complete finetuning in a zero-shot scenario. 

We also create XWINO --- a multilingual dataset of Winograd schemas that contains tasks from English, French, Japanese, Russian, Chinese, and Portuguese languages with the same specification.
We want to encourage research on commonsense reasoning in languages other than English and release our benchmark to facilitate the development and analysis of new methods for this problem.

Lastly, we demonstrate that the reasoning capabilities of cross-lingual models are concentrated in a small subset of attention heads located in higher layers of the model. Furthermore, this subset of heads is language-agnostic, which sheds light at another facet of linguistic universals encoded in models such as multilingual BERT and XLM-R.

\bibliographystyle{acl_natbib}
\bibliography{anthology,bibliography}

\clearpage
\appendix
\section{In-language metrics for supervised methods}
Here, we provide the metrics of our method and the finetuning baseline described by~\citet{kocijan-etal-2019-surprisingly} that were obtained on the training, validation and test data for the same language that the models were trained. Tables~\ref{tab:train_metrics_mbert} and~\ref{tab:train_metrics_xlmr} demonstrate the results: it can be seen that although our approach performs less well on the same language that was used for training, the issue of overfitting on train data is less noticeable, which might be the reason for better zero-shot metrics.

\begin{table}[h]
\centering
\small
\begin{tabular}{@{}llcc@{}}
\toprule
Model & Train lang & Train & Test \\ \midrule
\multirow{6}{*}{\citet{kocijan-etal-2019-surprisingly}} & en & 100.0±0.0 & 47.8±2.3 \\
 & fr & 100.0±0.0 & 44.4±7.9 \\
 & ja & 100.0±0.0 & 46.9±1.0 \\
 & ru & 100.0±0.0 & 52.5±1.4 \\
 & zh & 100.0±0.0 & 20.0±44.7 \\
 & pt & 100.0±0.0 & 49.6±6.2 \\ \midrule
\multirow{6}{*}{Ours} & en & 57.5±0.3 & 52.7±1.1 \\
 & fr & 54.5±3.2 & 33.3±17.2 \\
 & ja & 54.1±0.6 & 49.6±3.1 \\
 & ru & 60.8±0.9 & 46.2±2.3 \\
 & zh & 61.7±8.5 & 20.0±24.5 \\
 & pt & 57.1±0.4 & 43.0±6.9 \\ \bottomrule
\end{tabular}
\caption{Train and test set metrics for supervised methods, multilingual BERT.}
\label{tab:train_metrics_mbert}
\end{table}

\begin{table}[h]
\centering
\small
\begin{tabular}{@{}llcc@{}}
\toprule
Model & Train lang & Train & Test \\ \midrule
\multirow{6}{*}{\citet{kocijan-etal-2019-surprisingly}} & en & 100.0±0.0 & 83.3±1.0 \\
 & fr & 100.0±0.0 & 44.4±11.1 \\
 & ja & 100.0±0.0 & 79.6±2.2 \\
 & ru & 100.0±0.0 & 60.0±3.4 \\
 & zh & 100.0±0.0 & 50.0±0.0 \\
 & pt & 100.0±0.0 & 55.6±5.2 \\ \midrule
\multirow{6}{*}{Ours} & en & 71.6±0.4 & 67.2±1.2 \\
 & fr & 67.7±1.4 & 37.8±5.4 \\
 & ja & 71.2±0.2 & 65.6±1.7 \\
 & ru & 71.4±0.9 & 58.1±1.5 \\
 & zh & 73.3±3.3 & 20.0±24.5 \\
 & pt & 67.1±0.8 & 68.1±5.0 \\ \bottomrule
\end{tabular}
\caption{Train and test set metrics for supervised methods, XLM-R Large.}
\label{tab:train_metrics_xlmr}
\end{table}

\section{Analysis of common heads for multilingual BERT}
\label{app:analysis_mbert}
Table~\ref{tab:analysis-mbert} shows the evaluation results of models using top-5 attention heads of multilingual BERT. It can be seen that leaving only 5 heads out of 144 improves average accuracy in all cases and per-language accuracy in 18/30 cases without any significant decreases in quality.
\begin{table*}[h]
\centering
\begin{tabular}{@{}lccccccc@{}}
\toprule
Train lang          & Heads  & en             & fr             & ja             & ru             & zh             & Avg            \\ \midrule
\multicolumn{8}{c}{MAS (unsupervised)}                                                                                             \\ \midrule
\multirow{2}{*}{-}  & All    & 52.21          & 51.81          & 50.16          & 52.70          & \textbf{56.25} & 52.63          \\
 & Common & \textbf{56.60} & \textbf{53.01} & \textbf{51.82} & \textbf{60.00} & 50.00 & \textbf{54.29} \\ \midrule
\multicolumn{8}{c}{Ours (supervised)}                                                                                              \\ \midrule
\multirow{2}{*}{en} & All    & -              & 53.33          & 52.05          & 58.92          & 52.88          & 54.29          \\
                    & Common & -              & \textbf{54.53} & \textbf{52.52} & \textbf{59.78} & 52.25          & \textbf{54.77} \\ \cmidrule(l){2-8} 
\multirow{2}{*}{fr} & All    & 50.76          & -              & \textbf{50.41} & \textbf{51.80} & 50.06          & 50.76          \\
                    & Common & \textbf{51.01} & -              & 50.38          & 51.73          & \textbf{50.62} & \textbf{50.94} \\ \cmidrule(l){2-8} 
\multirow{2}{*}{ja} & All    & 53.25          & \textbf{52.64} & -              & 57.48          & \textbf{50.69} & 53.51          \\
                    & Common & \textbf{55.54} & 51.84          & -              & \textbf{58.51} & 50.56          & \textbf{54.12} \\ \cmidrule(l){2-8} 
\multirow{2}{*}{ru} & All    & 55.43          & 52.65          & \textbf{52.00} & -              & 49.62          & 52.43          \\
                    & Common & \textbf{56.20} & \textbf{52.92} & 51.66          & -              & \textbf{49.75} & \textbf{52.63} \\ \cmidrule(l){2-8} 
\multirow{2}{*}{zh} & All    & 50.28          & 50.12          & 50.09          & 50.53          & -              & 50.25          \\
 & Common & \textbf{50.82} & \textbf{50.14} & \textbf{50.24} & \textbf{51.30} & -     & \textbf{50.62} \\ \bottomrule
\end{tabular}
\caption{Performance of models trained with different sets of multilingual BERT attention heads.}
\label{tab:analysis-mbert}
\end{table*}

\section{Impact of number of heads for other languages}
\label{app:heads}

\begin{figure*}[t]
\centering
    \includegraphics[width=\textwidth]{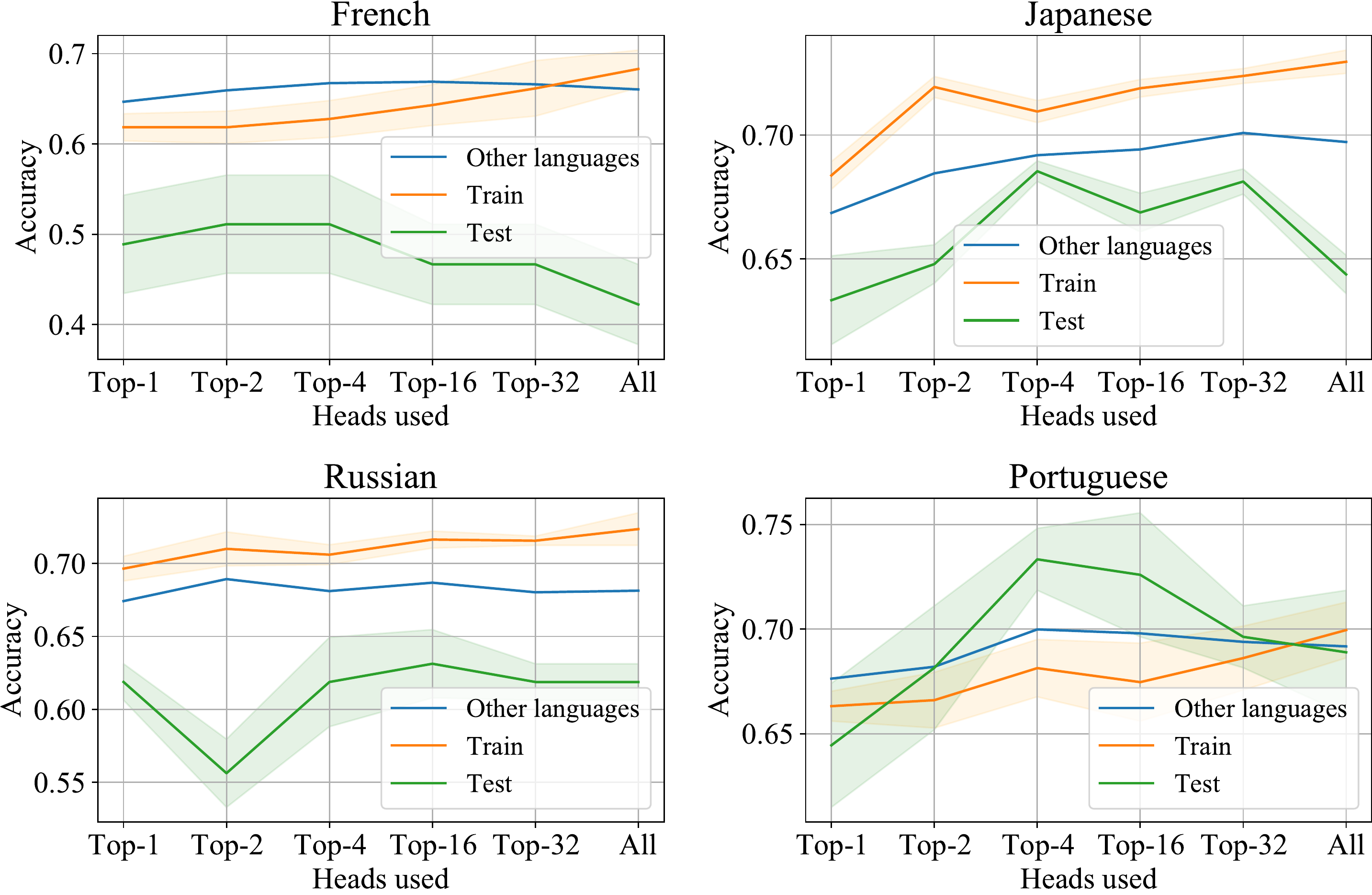}
    \caption{Effect of the number of used XLM-R attention heads on commonsense reasoning performance.}
\label{fig:heads_others}
\end{figure*}

In this section, we analyze the changes in both supervised and zero-shot performance for our method that follow from changes in the number of used attention heads. Figure~\ref{fig:heads_others} displays the results for French, Japanese, Russian, and Portuguese language; we omit the results for the Chinese language due to high variance from the small training dataset size. From this figure, we observe the same trend: increasing the number of used heads past 16 can favorably affect the accuracy on the training set, but negatively impacts the resulting quality both on the test set and for other languages.

\end{document}